\documentclass[conference]{IEEEtran}
\IEEEoverridecommandlockouts
\usepackage{booktabs}
\usepackage{makecell}
\usepackage{tabularray}
\usepackage{cite}
\usepackage{float}
\usepackage{amsmath,amssymb,amsfonts}

\usepackage{graphicx}
\usepackage{textcomp}
\usepackage{xcolor}
\def\BibTeX{{\rm B\kern-.05em{\sc i\kern-.025em b}\kern-.08em
    T\kern-.1667em\lower.7ex\hbox{E}\kern-.125emX}}

\usepackage{algorithm}
\usepackage{algpseudocode}

\begin{document}

\title{Prompt-Based Monte Carlo Tree Search for Mitigating Hallucinations in Large Models}

\author{\IEEEauthorblockN{ Zhihua Duan}
\IEEEauthorblockA{\textit{Intelligent Cloud Network Monitoring Department} \\
\textit{China Telecom Shanghai Company}\\
\textit{700 Daning Road, Shanghai, 200072}\\
Shanghai,China \\
duanzh.sh@chinatelecom.cn}
\and
\IEEEauthorblockN{ Jialin Wang*}
\IEEEauthorblockA{\textit{Executive Vice President} \\
\textit{Ferret Relationship Intelligence}\\
\textit{Burlingame, CA 94010, USA}\\
https://www.linkedin.com/in/starspacenlp \\
jialinwangspace@gmail.com}
 
}

\maketitle

\begin{abstract}
With the rapid development of large models in the field of artificial intelligence, how to enhance their application capabilities in handling complex problems in the field of scientific research remains a challenging problem to be solved. This study proposes an improved Monte Carlo Tree Search (MCTS) method based on prompt words. In the simulation search stage, it introduces dynamic adjustment of exploration parameters and adaptive selection strategies, which can better balance exploration and exploitation, thereby reducing the hallucination phenomenon. This paper takes the four subsets of the SciEval dataset as the test objects, and compares the Glm-4-flash+Improved MCTS method with the methods of several existing models. The results show that the Improved MCTS method performs better, providing new ideas and methods for the application of large models in the field of scientific research. 
\end{abstract}


\section{Introduction}
In the field of artificial intelligence, the development of Large Language Models (LLMs) is advancing at an unprecedented pace, with their capabilities in language understanding and generation continuously improving. However, effectively enhancing the application capabilities of large models in the field of scientific research, especially when dealing with complex problems, remains a challenge. To address this challenge, this study proposes an improved Monte Carlo Tree Search (MCTS) algorithm based on prompt words. By dynamically adjusting exploration parameters and adaptively selecting simulation strategies, the efficiency of LLMs in solving scientific problems is enhanced.

The Monte Carlo Tree Search (MCTS) algorithm is a combination of the Monte Carlo method and tree search, widely used in decision-making processes and game problems. Through four steps of selection, expansion, simulation, and backpropagation, it explores and evaluates the decision tree to find the optimal action plan. This study improves upon the existing MCTS by introducing a function for dynamically adjusting exploration parameters and a function for adaptively selecting simulation strategies during the simulation search phase. These functions determine whether to use a greedy strategy or a random strategy based on the complexity of the problem and preset thresholds.

To evaluate the improved MCTS algorithm, we employed the SciEval dataset, a comprehensive and multi-dimensional evaluation benchmark specifically designed for assessing the capabilities of large language models in the field of scientific research. The SciEval dataset covers dimensions such as basic knowledge, knowledge application, scientific computation, and research capabilities, combining diverse data sources and rich data types to provide a comprehensive, reliable, and challenging tool for evaluation. This study tested four subsets of SciEval and compared the performance of the Glm-4-flash+Improved MCTS strategy with existing models such as GPT-3.5-Turbo+CoT, GPT-3.6-Turbo+ToT, and GPT-3.7-Turbo+ReST-MCTS. Through testing on the four subsets of SciEval, we found that the improved MCTS strategy achieved an average score of 65.6, demonstrating superior performance.

By introducing dynamic adjustment of exploration parameters and adaptive selection of simulation strategies, this study has improved the MCTS algorithm, providing a new perspective for the application of LLMs in the field of scientific research.
 
\section{Related Work}
\subsection{Chain of Thought Prompt}
Propose a chain of thought method that uses Scratchpads for multi-step calculations\cite{showwork}.Enhance complex reasoning abilities by generating a chain of thought prompts consisting of a series of intermediate reasoning steps\cite{chain}."Let's think step by step," can demonstrate good zero-shot reasoning ability\cite{Large}.Selecting more complex reasoning chains and prompts with a greater number of reasoning steps improves the performance on multi-step reasoning tasks\cite{Complexity}.

\subsection{Monte Carlo Tree Search (MCTS) method}
Positioning large models as world models and reasoning agents, the planning algorithm based on Monte Carlo Tree Search (MCTS) has achieved efficient solutions to complex reasoning problems\cite{Reasoning}.The Tree of Thoughts (ToT) has been proposed, which enables large models to explore different reasoning paths and self-assess for selection\cite{tree}.A reasoning method called rStar is proposed, which can significantly enhance the reasoning capabilities of small models\cite{Mutual}.An innovative algorithm called MCT Self-Refine (MCTSr) has been proposed\cite{Accessing}.

\subsection{Post-training enhancement of reasoning capabilities}
Some studies employ supervised fine-tuning or reinforcement learning methods to train large models.The ReST-MCTS method is a reinforcement self-training approach that combines process reward guidance with tree search Monte Carlo Tree Search (MCTS)\cite{ReST}.Using the Rejection Sampling Fine-Tuning (RFT) method effectively enhances mathematical reasoning performance\cite{Scaling}.
Training with the modified DPO loss and the negative log-likelihood term\cite{Iterative}.Propose the ReFT (Reinforcement Fine-Tuning) method to enhance the reasoning capabilities of large models through reinforcement fine-tuning\cite{Proceedings}.

Unlike the aforementioned methods, we propose an improved Monte Carlo Tree Search (MCTS) method, which can significantly enhance the model's reasoning capabilities.

\section{Methods}
As shown in Figure 1, it is an improved Monte Carlo method based on prompt words. The Monte Carlo Tree Search (MCTS) algorithm is a combination of the Monte Carlo method and tree search, mainly consisting of four steps: selection, expansion, simulation, and backpropagation.

\begin{figure*}[htbp]
  \centering
  \includegraphics[width=\linewidth]{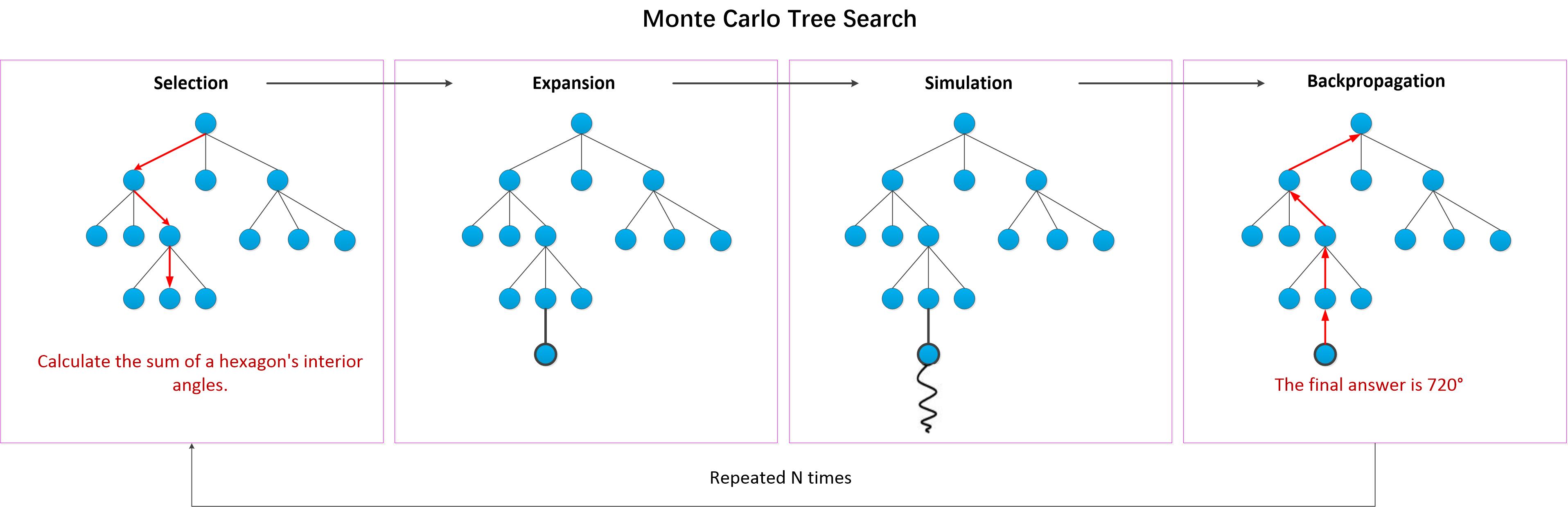}
  \caption{ Schematic diagram of the full process of the Monte Carlo Tree Search (MCTS) method.} 
\end{figure*}

\begin{itemize}
  \item \textbf{Selection}: Starting from the root node, recursively select the optimal child node until a leaf node is reached.
  \item \textbf{Expansion}: If the current leaf node is not a terminal node, then create one or more child nodes and select one of them for expansion.
  \item \textbf{Simulation}: Starting from the expanded node, run a simulated outcome until the end of the game.
  \item \textbf{Backpropagation}: Propagate the simulation results back to the root node of the tree, updating the information of each node along the path, including the number of visits and node value.
\end{itemize}

By iteratively cycling through these four steps, the MCTS algorithm can effectively explore and evaluate the decision tree to find the optimal action plan.

This study makes improvements based on existing Monte Carlo work: during the simulation search phase, a function is dynamically adjusted to reduce the exploration parameter according to the number of visits. An adaptive function for selecting simulation strategies is used, which decides whether to employ a greedy strategy or a random strategy based on the complexity of the problem and a preset threshold.
Dynamic adjustment of exploration parameters: A function for dynamically adjusting exploration parameters is defined, which adjusts the exploration parameters dynamically according to the number of visits to a node. In the early stages of exploration, when the number of visits to a node is low, the exploration parameter is larger, leading to more exploration and trying out different paths to avoid prematurely falling into local optimal solutions. As exploration progresses and the number of visits to a node increases, the exploration parameter gradually decreases, causing the algorithm to lean more towards utilizing existing information and selecting more promising paths for in-depth exploration, thereby more flexibly balancing exploration and improving search efficiency.
Adaptive selection of simulation strategies: A Monte Carlo Tree Search (MCTS) algorithm that combines problem complexity assessment and adaptive simulation strategy selection is proposed. A function complexity(problem) is designed to assess the complexity based on the length of the problem. Through the function choose simulation policy(mcts task, node), which selects simulation strategies adaptively, the MCTS algorithm decides whether to use a greedy strategy or a random strategy for simulation based on the complexity of the problem at the current node and a preset threshold. The MCTS algorithm can adaptively select simulation strategies according to the complexity of the problem, thereby improving efficiency on different problems.

\section{Experimental Design}

\subsection{Dataset}
The SciEval dataset is a comprehensive and multi - dimensional evaluation benchmark designed specifically for the capabilities of large - scale language models in the field of scientific research. By covering four major dimensions of basic knowledge, knowledge application, scientific computing, and research ability, and combining diverse data sources and rich data types, including both objective and subjective questions, it provides a comprehensive, reliable, and challenging evaluation tool for the application capabilities of large - scale language models in the scientific field. This study tested four subsets of SciEval: part1, part2, part3, and part4.
\subsection{Experimental Results}
As shown in Table I, this study compared the performance of GPT-3.5-Turbo+CoT, GPT-3.5-Turbo+ToT, GPT-3.5-Turbo+ReST-MCTS, and Glm-4-flash+Improved MCTS on the four subsets of SciEval. The CoT (Chain of Thought) strategy is a method of solving problems through step-by-step reasoning. The ToT (Tree of Thoughts) strategy is a structured method for problem-solving and reasoning, which decomposes problems into multiple branches and levels, explores possible solutions step by step, and ultimately selects the optimal path. It is capable of handling complex decision-making and multi-step reasoning tasks. The ReST-MCTS (Reinforcement Learning with Monte Carlo Tree Search) strategy combines reinforcement learning and Monte Carlo tree search techniques.

\begin{table}
\centering
\caption{Comparison of Model Performance on SciEval Subsets}
\begin{tblr}{
  width = \linewidth,
  colspec = {Q[219]Q[244]Q[90]Q[90]Q[90]Q[90]Q[90]},
  hlines,
}
Models        & Method        & Part1 & Part2 & Part3 & Part4 & Ave   \\
GPT-3.5-Turbo$^{\dag}$ & CoT           & 28.88 & 78.61 & 70.00    & 71.46 & 62.24 \\
GPT-3.5-Turbo$^{\dag}$ & ToT           & 32.50  & 76.11 & 68.05 & 67.59 & 61.06 \\
GPT-3.5-Turbo$^{\dag}$ & ReST-MCTS     & 29.72 & 78.88 & 69.72 & 70.91 & 62.31 \\
Glm-4-flash   & Improved-MCTS & 53.3  & 82.70  & 67.70  & 58.72 & 65.60  
\end{tblr}
\footnotesize{ Scores with $^{\dag}$ were obtained from the paper\cite{ReST}}\\
\end{table}

This paper is based on the evaluation of the Glm-4-lash large - scale model. Glm-4-lash performs well in real - time web search, long - context processing, and multilingual support, making it suitable for a variety of application scenarios such as intelligent question - answering, summary generation, and text data processing. This paper adopts an improved Glm-4-lash+MCTS (Monte Carlo Tree Search) strategy, which is an improved Monte Carlo method. The experimental results show that the improved MCTS strategy has an average score of 65.6 on the four subsets of SciEval, outperforming existing models such as GPT - 3.5 - Turbo+CoT, GPT - 3.5 - Turbo+ToT, and GPT - 3.5 - Turbo+ReST - MCTS.

\subsection{Case Study} 
As shown in Figure 2, by exploring the calculation of the sum of the interior angles of a hexagon from the perspective of the Monte Carlo method, it demonstrates the approach of the Monte Carlo method in dealing with mathematical problems. A user poses a question: "Calculate the sum of the interior angles of a hexagon." In the first step, three paths are attempted, each describing a method for calculating the sum of the interior angles of a hexagon. In the second step, an expansion is made on Path 1, through simulation and backtracking calculations, and ultimately, the sum of the interior angles of the hexagon is calculated to be 720 degrees. For a simple problem like the sum of the interior angles of a hexagon, which has a clear formula, thinking and trying from different angles using the Monte Carlo method can also bring new insights and a deeper understanding of the problem.

\begin{figure*}[htbp]
  \centering
  \includegraphics[width=\linewidth]{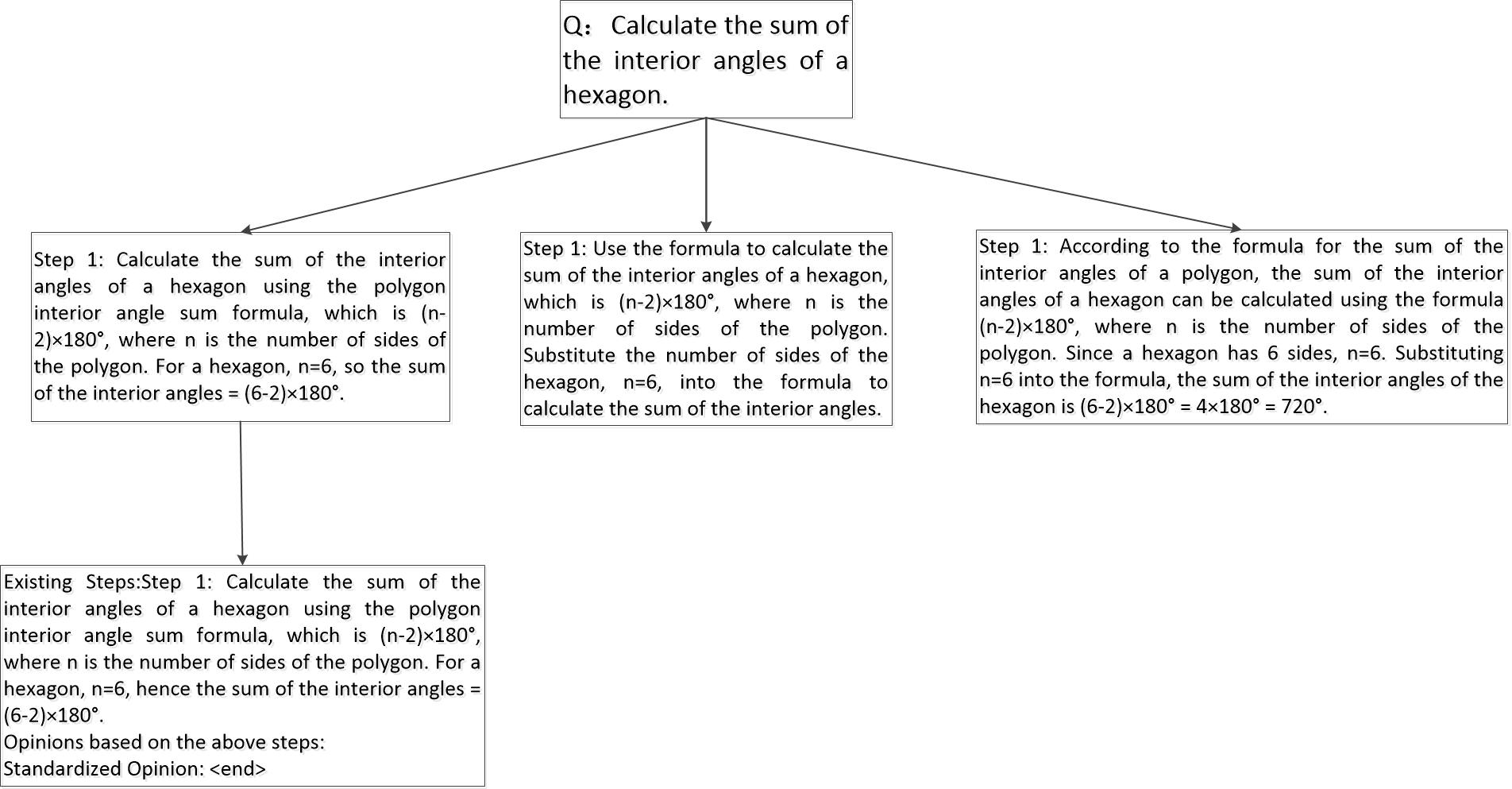}
  \caption{Calculate the sum of the interior angles of a hexagon.} 
\end{figure*}

\section{Limitations and Future Work}
Although this study has made some progress in enhancing the application capabilities of LLMs in the field of scientific research, there are still some potential directions for improvement:
1.Further optimize the dynamic adjustment mechanism: More complex dynamic adjustment functions can be explored to better adapt to different types of scientific problems.
2.Expand the evaluation dataset: In addition to the SciEval dataset, more scientific domain datasets can be considered to comprehensively evaluate the generalization ability of the improved MCTS algorithm.
3.Combine with other optimization techniques: The improved MCTS algorithm can be combined with other optimization techniques (such as reinforcement learning) to further enhance the performance of reasoning.

\subsection{ Use LangGraph to enhance the Improved MCTS}
LangGraph can be used to enhance the improved Monte Carlo Tree Search (MCTS).

1. Integration of Cyclical Workflows

The MCTS process naturally involves iterative cycles of selection, expansion, simulation, and backpropagation. LangGraph's core strength in supporting cycles and branching makes it an ideal framework for implementing MCTS workflows.
Benefit: Streamlined development of the iterative process, ensuring consistency in handling the dynamic adjustments and adaptive selection strategies described in the paper.

2. State Persistence and Recovery

LangGraph's built-in persistence can save the state of MCTS at each node, including intermediate scores, exploration parameters, and simulation strategies.
Application: If the process is interrupted (e.g., due to resource constraints or the need for human intervention), LangGraph allows seamless resumption without reinitializing the MCTS workflow.
Example: When dynamically adjusting exploration parameters, persistent states ensure that transitions between high exploration and exploitation phases are reliably managed.

3. Human-in-the-Loop for Dynamic Adjustments

The paper emphasizes the use of adaptive simulation strategies based on problem complexity. LangGraph enables human oversight to validate or refine these strategies in real time.
Use Case: A human reviewer can intervene during simulation or backpropagation to verify results or adjust the thresholds for exploration parameters.

4. Improved Controllability

The fine-grained control provided by LangGraph allows for dynamic adjustments to exploration parameters and simulation strategies, as described in the paper.
Example: Implementing the function complexity(problem) and the adaptive simulation policy choose\_simulation\_policy(mcts\_task, node) can be directly mapped as nodes in a LangGraph workflow.

\section{CONCLUSION}
This study proposes an improved Monte Carlo Tree Search (MCTS) algorithm based on prompt words, aiming to enhance the application capabilities of Large Language Models (LLMs) in the field of scientific research, especially in terms of efficiency when dealing with complex problems. By dynamically adjusting exploration parameters and adaptively selecting simulation strategies, the improved MCTS algorithm can more flexibly balance exploration and exploitation during the simulation search phase, providing new perspectives and methods for the application of LLMs in the field of scientific research. Although this study has made some progress in enhancing the application capabilities of LLMs in the field of scientific research, there are still some potential directions for improvement. Future work will continue to explore and optimize these methods to further improve the model's performance in solving complex scientific problems. 
 
\section{Acknowledgement}
The authors express their gratitude to GLM-4-Flash from Zhipu AI Technology Co., Ltd. in Beijing, China, for their assistance with large language models.


\end{document}